# Environmental Sound Recognition using Masked Conditional Neural Networks

Fady Medhat    David Chesmore    John Robinson

Department of Electronic Engineering
University of York, York
United Kingdom
`{fady.medhat,david.chesmore,john.robinson}@york.ac.uk`

**Abstract.** Neural network based architectures used for sound recognition are usually adapted from other application domains, which may not harness sound related properties. The ConditionaL Neural Network (CLNN) is designed to consider the relational properties across frames in a temporal signal, and its extension the Masked ConditionaL Neural Network (MCLNN)[1] embeds a filterbank behavior within the network, which enforces the network to learn in frequency bands rather than bins. Additionally, it automates the exploration of different feature combinations analogous to handcrafting the optimum combination of features for a recognition task. We applied the MCLNN to the environmental sounds of the ESC-10 dataset. The MCLNN achieved competitive accuracies compared to state-of-the-art convolutional neural networks and handcrafted attempts.

## 1      Introduction

Handcrafting the features required for the sound recognition problem has been investigated for decades. It is still an open area of research that consumes a lot of effort in an attempt to design the best features to be fed to a recognition system. Deep architectures of neural networks are currently being considered to be a replacement to the feature handcrafting stage. There is an endeavor to use these deep architectures that achieved wide success for images [1], on signals like sound to extract features automatically that can be further classified using a conventional classifier such as a Support Vector Machine (SVM) [2]. Handcrafted features still hold their position, but the performance gap between them and the use of the deep neural architectures as automatic feature extractors is getting narrower.

---

[1] Code: https://github.com/fadymedhat/MCLNN

[†]This work is funded by the European Union's Seventh Framework Programme for research, technological development and demonstration under grant agreement no. 608014 (CAPACITIE).

The work by Soltau et al. [3] marks an early attempt in using neural networks based architectures for feature extraction in sound recognition. In their work, they used a three-stage recognition system. The first stage is an event detection phase to extract musical events, where they dropped the output nodes of a neural network and used the hidden nodes as features for the succeeding stages. A more recent attempt by Lee et al. [4] used Convolutional Deep Belief Networks [5] for several audio recognition tasks. Hamel et al. [6] introduced another attempt to extract features using a Deep Belief Network (DBN) architecture, where the features extracted with the DBN were further classified using an SVM, a similar attempt was in [7]. Hinton et al. [8] proposed the use of deep neural network architectures to replace the Gaussian Mixture Model (GMM) in a GMM-HMM combination for speech recognition. Graves et al. [9] used a deep recurrent neural network for speech recognition. An overlapping usage to music genre classification was in the work of Oord et al. [10], where they used Convolutional Neural Network (CNN) [11] for automatic music recommendation. Dieleman et al. [12] aimed to bypass the need for an intermediate signal representation like spectrograms by using a CNN over the raw signal, where their findings showed that spectrograms provided better performance.

Several neural based architectures were proposed for the sound recognition problem in music, speech and environmental sound, but they are usually adapted to the sound problem after they gain wide acceptance by the research community in other applications especially image recognition. Despite these successful attempts, they may not harness the full properties of the sound signal represented in a spectrogram. For example, recent efforts [13, 14] proposed the need to restructure the widely used Convolutional Neural Network (CNN) to fit the sound recognition problem. This need is attributed to the inability of the vanilla CNN to preserve the spatial locality of the learned features across the frequency domain in a time-frequency representation. Similarly, DBNs have a shortcoming in ignoring the inter-frames relation, where it treats each frame as an isolated entity.

The ConditionaL Neural Network (CLNN) [15] is designed to exploit the time-frequency representation of the sound signals. The model structure also extends its application to other multi-dimensional temporal representations. The CLNN takes into consideration the influence the successive frames, in a temporal signal, have on each other. The Masked Conditional Neural Network (MCLNN) [15] extends the functionality of the CLNN to include a binary masking operation using a systematic sparseness to automate the exploration of different feature combinations concurrently, which is usually a manual mix-and-match process of various features combinations. Additionally, the MCLNN embeds a filterbank behavior, which lends filterbank's properties such as the frequency shift-invariance to the neural network structure. In this work, we extend the work in [16] with an analysis of the masking effect in the MCLNN compared to the CLNN.

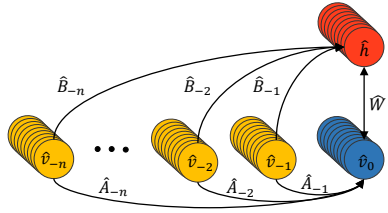

Fig. 1. The Conditional RBM structure

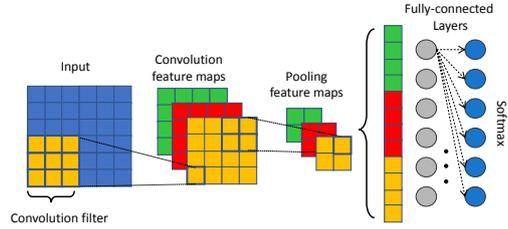

Fig. 2. Convolutional Neural Network.

## 2   Related Work

The Conditional Restricted Boltzmann Machine (CRBM) [17] is an extension to the RBM [18] that takes into consideration the temporal nature of the successive feature vectors. The CRBM has been introduced initially to model the human motion by training it on a temporal signal captured from the movement of the human joints.

Fig. 1 shows the structure of a CRBM. The main difference compared to the RBM involves the inclusion of the directed links from previous visible frames ($\hat{v}_{-n}, ..., \hat{v}_{-2}, \hat{v}_{-1}$) to both the visible feature vector $\hat{v}_0$ and the hidden nodes $\hat{h}$. The figure depicts the $\hat{W}$ links present in a normal RBM to hold the bidirectional relation between the hidden and the visible nodes. Additionally, the $\hat{B}$ links to hold the conditional relation between the previous *n* states of the visible vectors and the current hidden one, and the $\hat{A}$ links to hold the autoregressive relations between the previous visible states and the current visible vector $\hat{v}_0$. The Interpolating CRBM (ICRBM) [19] extended the work of the CRBM by considering both the future frames in addition to the past ones used in the CRBM. The ICRBM was applied for the phone recognition task in [19], and it surpassed the accuracy of the CRBM. The CLNN discussed in the next section extends from both the CRBM and the ICRBM by considering both the future and past frames without the autoregressive links.

The Convolutional Neural Network (CNN) [11] achieved breakthrough results [1, 20] for the image recognition problem. The model is based on two primary operations namely; Convolution and Pooling. The convolution operation involves a set of filters (weight matrices) of small sizes, e.g. 5×5, to convolve the 2-dimensional input image as shown in Fig. 2. The pooling (e.g. mean or max pooling) operation is a subsampling procedure to decrease the resolution of the feature maps generated from the convolution stage. The feature maps generated from these interleaved convolution and pooling operations are flattened to a single feature vector to be fed to a fully-connected network of neurons or a conventional classifier, e.g. Random Forest, SVM for the final classification decision.

Long Short-Term Memory [21] is a Recurrent Neural Network (RNN) model that makes use of internal memory to capture the previous states influence on the current input. Choi et al. [22] tried a hybrid model of both the CNN to extract localized fea-tures and the LSTM to capture long-term dependencies in the Convolutional RNN (CRNN) for several music tasks.

Convolutional DBN (ConvDBN) [5] by Lee et al. extended the CNN terminologies like weight sharing and pooling to the unsupervised learning of the DBN. In their work, they adopted a convolution layer like the one in the CNN to the ConvDBN in addition to a probabilistic max-pooling layer. The ConvDBN was used in [4] for speech and music recognition.

The deep neural network architectures proposed earlier are examples of the most successful attempts, but they are adapted to the sound recognition problem after they gain wide acceptance in other domains primarily image recognition. This may not optimally harness properties of multidimensional temporal signals. For example, convolutional models dependent on weight sharing, which does not preserve the spatial locality of the learned features and models such as DBN treats each temporal frame as an isolated entity, ignoring the inter-frames relation.

The ConditionaL Neural Networks (CLNN) considers the inter-frame relation, and the Masked ConditionaL Neural Networks extends the CLNN architecture by embedding a filterbank-like behavior within the network. This enforces the network to learn about frequency bands allowing it to sustain frequency shifts as in a normal filterbank. Additionally, the masking used in the MCLNN automates the exploration of different feature combination similar to the manual process of finding the optimum combination of features.

## 3    Conditional Neural Networks

The ConditionaL Neural Network (CLNN) [23] is a discriminative model that stems from the generative CRBM by considering the conditional links between the previous visible state vectors and the hidden nodes. The CLNN also considers the future frames in addition to the previous ones as in the ICRBM.

For notation purposes, multiplication is denoted by the ( $\times$ ) or the absence of any sign between terms, e.g. ( $xW$ ) or ( $l \times e$ ). Element-wise multiplication between vectors or matrices of the same sizes is denoted by ( $\circ$ ). The hat operator is used to denote vectors when used in combination with lowercase literals ( $\hat{x}$ ) and to denote matrices with uppercase ones ( $\widehat{W}$ ). Subscripts in the absence of the hat operator are used to indicate individual elements in a vector or a matrix e.g. $x_i$ is the $i^{th}$ element in vector ( $\hat{x}$ ) and $W_{i,j}$ is the element at position [$i, j$] within a matrix. ( $\widehat{W}_u$ ) denotes a matrix at index $u$ within a tensor.

The CLNN is formed of a vector-shaped hidden layer of $e$ dimensions and accepts an input of a dimension [$l, d$], where $l$ is the feature vector length and $d$ is the number of frames in a window of width following (1)

$$d = 2n + 1 \ , n \geq 1 \qquad (1)$$

where $d$ is the window's width and $n$ is the order. The order $n$ refers to the number of frames to consider in a single temporal direction. The 2 is for both the past and future frames, and the central frame is considered by the 1. Each feature vector within the window of frames is fully-connected to the same hidden layer of length *e* neurons. The activation of a single hidden node can be given in (2)

$$y_{j,t} = f\left(b_j + \sum_{u=-n}^{n} \sum_{i=1}^{l} x_{i,u+t} \, W_{i,j,u}\right) \quad (2)$$

where $y_{j,t}$ is the activation at the $j^{th}$ hidden neuron (index *t* discussed later) and *f* is the transfer function applied at the neuron level. $b_j$ is the bias at the $j^{th}$ neuron. $x_{i,u+t}$ is $i^{th}$ element of the vector of length *l* at index *u+t*, where *u* is the index of the feature vector in a window of width *d*, i.e. *u* takes the values within the interval [-*n*, *n*]. $W_{i,j,u}$ is the weight between the $i^{th}$ feature in the input vector at index *u* and the $j^{th}$ hidden neuron. The window of frames is extracted from a larger chunk of the spectrogram, which we will refer to as the segment. The index *t* in the above equation refers to the position of the frame within the segment, which is also the window's central frame (at *u*=0). The window's additional 2*n* frames are also extracted from the segment together with the window's central frame at index *t*. The vector form of (2) is formulated in (3)

$$\hat{y}_t = f\left(\hat{b} + \sum_{u=-n}^{n} \hat{x}_{u+t} \cdot \widehat{W}_u\right) \quad (3)$$

where the activation vector $\hat{y}_t$ for the vector $\hat{x}_t$ condition on the 2*n* vectors in the window of frames is given by the output of the transfer function *f*. The bias vector of the hidden layer is $\hat{b}$. $\hat{x}_{u+t}$ is the input feature vector at index $u + t$ within the window, where *t* is the window's middle frame and also the index of the frame in the segment. The index *u* is used to specify the position of the input frame in the window, where the middle frame is positioned at $u = 0$. $\widehat{W}_u$ is the weight matrix at position *u* within the weight tensor of size [feature vector length *l*, hidden layer width *e*, window's depth *d*]. The vector-matrix multiplication between the vector of length *l* at index *u* in a window and its corresponding weight matrix at index *u* within the weight tensor, generates *d* frames summed together per dimension to produce a single resultant vector of *e* dimensions. A logistic transfer function provides the conditional distribution of the prediction of the window's middle frame conditioned on the 2*n* neighboring frames formulated in $p(\hat{y}_t | \hat{x}_{-n+t}, \ldots \hat{x}_{-1+t}, \hat{x}_t, \hat{x}_{1+t}, \ldots \hat{x}_{n+t}) = \sigma(\ldots)$, where $\sigma$ is a Sigmoid or the final layer output Softmax function.

The output of a CLNN layer is 2*n* fewer frames than its input. Therefore, in a deep CLNN architecture, the input segment should account for the consumed frames at each layer. Accordingly, the width of a segment of frames is given in (4)

$$q = (2n)m + k \quad , n, m \text{ and } k \geq 1 \quad (4)$$

where the segment of width *q* (the length follows the feature vector length *l* ) is given by the order *n*, the number of layers *m* and the extra frames *k*. The extra frames *k* account for the frames that should remain beyond the CLNN layers to be flattened to a single vector or globally pooled as in [24] but for time-frequency representations, it is a single dimensional pooling.

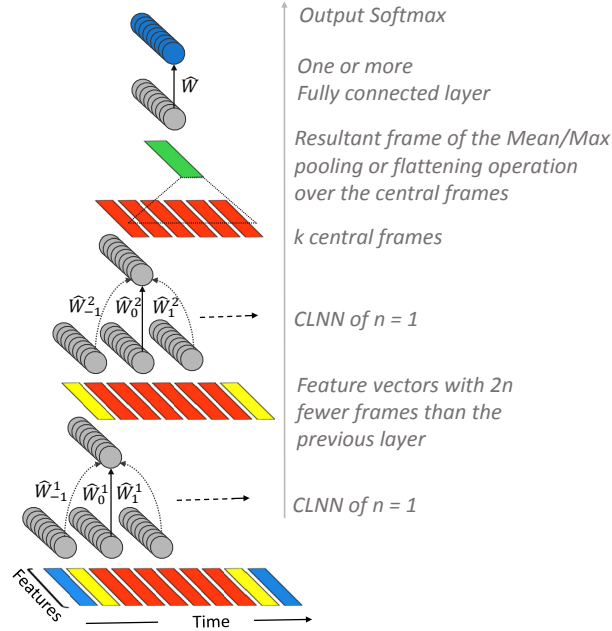

**Fig. 3.** A two-layered CLNN model

Fig. 3 shows a two-layer CLNN model. Each CLNN layer has an order $n = 1$, accordingly, a window considers one past and one future frame in addition to the central one. Each CLNN layer possesses a weight tensor $\widehat{W}^b$, where $b$ (the layer index)=1, 2,…, $m$. The number of matrices in the tensor matches the number of frames in the window. Accordingly, at $n=1$, $\widehat{W}_0^b$ will processes the central frame, $\widehat{W}_{-1}^b$ processes the previous frame and $\widehat{W}_1^b$ processes one future frame. The second CLNN operates on the frames from the first CLNN layer. The figure also depicts a number of frames remaining beyond the two CLNN layers used in the flattening or pooling operation, which behaves as a form of aggregation over a texture window studied in [25] for music.

## 4     Masked Conditional Neural Networks

Time-frequency representations such as spectrograms are widely adapted for signal analysis. These intermediate representations provide an insight of the change of energy across different frequency components as the signal progresses through time. Though these representations are efficient, they are sensitive to frequency shifts, especially for sound, since the same sound signal can be affected by different propagation factors that can alter its spectrogram representation. This is encountered through the use of a filterbank, which is one of the central operating blocks in a Mel-Scaled transformation such as an MFCC. A filterbank is a group of filters with their central frequency spaced from each other following a specific distance, e.g. Mel-spaced in MFCC. The filters allow each group of frequency bins to be treated as a band of frequencies, which permits the spectrogram transformation to be frequency shift-invariant.

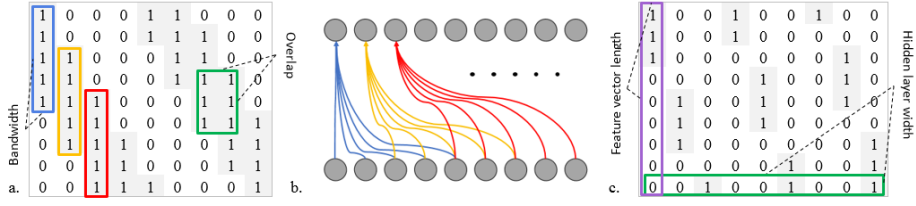

**Fig. 4.** Examples of the Mask patterns. a) A bandwidth of 5 with an overlap of 3, b) The allowed connections matching the mask in a. across the neurons of two layers, c) A bandwidth of 3 and an overlap of -1

The Masked ConditionaL Neural Network (MCLNN) [23] extends the CLNN by adopting a filterbank-like behavior within the network by enforcing a systematic sparseness through a binary mask. The binary mask is designed using two parameters: the Bandwidth *bw* and the Overlap *ov*. The bandwidth specifies the number of consecutive ones, column-wise, as shown in Fig. 4.a. and the overlap specifies the superposition of the ones between one column and another. This binary mask, when enforced over the network connections, allows each hidden node to be an expert in a localized region of the feature vector. Thus, preserving the spatial locality of the features especially for time-frequency representations. Fig. 4.b. shows the sparseness pattern enforced on the links matching the mask in Fig. 4.a (ignoring the temporal aspect for simplicity). The linear indices of the 1's locations to generate a masking pattern follows (5)

$$lx = a + (g-1)(l + (bw - ov)) \quad (5)$$

where the linear index *lx* of a 1's positions is dependent on the feature vector length, the bandwidth *bw* and the overlap *ov*. The values of *a* range within the interval [0, *bw*-1] and *g* within interval $[1, \lceil (l \times e)/(l + (bw - ov)) \rceil ]$.

The overlap can be assigned negative values as depicted in Fig. 4.c., which refers to the non-overlapping distance between groupings of ones across columns. The figure also shows the exploration of different feature combinations concurrently within the network. This is depicted in the presence of the shifted version of the filterbank-like pattern across the 1st set of three columns compared to the 2nd and the 3rd sets. In this setting, the input to the 1st neuron (mapping to the first column) is the first three features, the 4th neuron is observing the first two features, and the 7th neuron has an interest in the first feature only. This behavior enforces different hidden nodes to consider a different combination of features through disabling the effect of various regions of the input and enabling others.

The mask is applied through an element-wise multiplication between the mask pattern and each matrix within the weight tensor following (6)

$$\hat{Z}_u = \hat{W}_u \circ \hat{M} \tag{6}$$

where $\hat{W}_u$ is the original weight matrix at index $u$ masked by the mask $\hat{M}$. The original weight matrix in (3) is substituted with the masked version $\hat{Z}_u$.

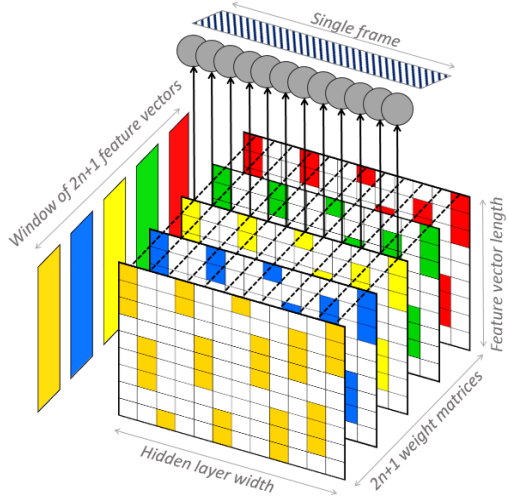

**Fig. 5.** A single step of the MCLNN

Fig. 5 shows an MCLNN step. The figure depicts the number of matrices matching the number of frames. The highlighted regions in each matrix represent the active connection following the binary mask pattern. The output generated from each of the vector-matrix multiplication is $d$ vectors of $e$ dimensions each. The vectors are summed together feature-wise to generate one representative vector for the window then the nonlinearity is applied through the transfer function over this single vector.

## 5   Experiments

We extend the evaluation of the MCLNN in [16] to different experimental settings affecting the model performance. We evaluated the MCLNN using the ESC-10 dataset of environmental sounds.

The ESC-10 [26] is a dataset of 10 environmental sounds categories evenly pre-distributed into 5-folds with 40 samples per category: Dog Bark, Rain, Sea Waves, Baby Cry, Clock Tick, Person Sneeze, Helicopter, Chainsaw, Rooster and Fire Cracking. The dataset is released with all files unified to 5 seconds in length, with clips having shorter events padded with silence. As an initial pre-processing step, we trimmed the silence and cloned the files several times to extract 5 seconds.

**Table 1.** MCLNN model parameters for ESC-10

| Layer | Type | Nodes | Mask Bandwidth | Mask Overlap | Order $n$ |
|---|---|---|---|---|---|
| 1 | MCLNN | 300 | 20 | -5 | 15 |
| 2 | MCLNN | 200 | 5 | 3 | 15 |

All files were transformed to a 60 bins logarithmic Mel-scaled spectrogram with the Delta (1st derivative between the frames) concatenated column-wise to generate 120 frequency bins vector. The transformation used an FFT window of 1024 and 50% percent overlap.

We adopted a two-layer MCLNN architecture followed by a single dimension global mean pooling [24] layer to pool across the extra frames $k$ and two fully connected layers of 100 nodes before the final Softmax. Parametric Rectifier Linear Units (PRELU) [27] were used as the transfer function for all the model's nodes. Dropout [28] was used for regularization. A summary of the hyper-parameters used is given in Table 1. The whole model was trained using ADAM [29] to minimize the categorical cross-entropy, and the final decision of a clip's class is decided by a probability voting across the frames of the clip following (7).

$$Category = argmax_{j=1...c} \left( \sum_{i=1}^{r} o_{ji} \right) \qquad (7)$$

where each clip has a number of prediction vectors $r$ matching the number of segments. The vector has $c$ elements corresponding to the number of classes. All activation vectors are summed, and the class with the maximum value is the clip's category.

The mask overlap and bandwidth are different across the two MCLNN layers as listed in Table 1. We found through several experiments that using wide bandwidth with negative overlap in the first layer in combination with narrow bandwidth and positive overlap in the second layer increases the accuracy. This is accounted to the ability of the wide bandwidth to collectively consider more frequency bins, which suppress the effect of the smearing of the energy. Also, the negative overlap increases the sparseness, which decreases the effect of noisy bins in the input. On the other hand, the second layer's narrow bandwidth and positive overlap allow considering a smaller number of bins collectively as a band, which focuses on the distinct features that can enhance the accuracy.

Experiments were applied using cross-validation over the dataset's original 5-fold, where 3 training folds are standardized, and the mean and the standard deviation of the training data were used to standardize the testing and validation folds.

Table 2 lists several accuracies achieved using both the CLNN and the MCLNN. MCLNN achieved an accuracy of 85.5% at k=40 and an accuracy of 83% at k=1. The deep CNN architecture used in Piczak-CNN [30] is a model of two convolutions and two pooling layers followed by two densely connected layers of neurons of 5000 neurons each. Piczak-CNN experimented with different segment sizes, where he used short segments of 41 frames and long segments formed of 101 frames. Piczak reported higher accuracy using the long segments compared to the short segments. Piczak-

**Table 2.** Accuracies reported on the ESC-10

| Classifiers and features | Acc.% |
| --- | --- |
| **MCLNN ($k$=40) + Mel-Spectrogram (this work)[1]** | **85.50** |
| MCLNN ($k$=1) + Mel-Spectrogram [16][1] | 83.00 |
| **MCLNN ($k$=25) + Mel-Spectrogram (this work)[1]** | **82.00** |
| Piczak-CNN + Mel-Spectrogram (*101 frames*) [30][2] | 80.00 |
| Piczak-CNN + Mel-Spectrogram (*41 frames*)[30][2] | 78.20 |
| **CLNN ($k$=25) + Mel-Spectrogram (this work)[1]** | **77.50** |
| **CLNN ($k$=40) + Mel-Spectrogram (this work)[1]** | **75.75** |
| **CLNN ($k$=1) + Mel-Spectrogram (this work)[1]** | **73.25** |
| Random Forest + MFCC [26][1] | 72.70 |

[1] *Without augmentation*
[2] *With augmentation*

CNN achieved 80% using 25 million parameters with the long segment of 101 frames. Additionally, the work of Piczak [30] applied 10 augmentation variants for each sound file, which involves applying 10 deformations to the sound file with different time delays and shifted pitches. This increases the size of the dataset and consequently the accuracy as studied by Salamon et al. in [31]. On the other hand, MCLNN achieved the listed accuracies using 3 million parameters (12% of the parameters used in the Piczak-CNN) without any augmentation. MCLNN achieved 85.5% for long segments containing 101 frames ($m$=2, $n$=15, $k$=40 and middle frame. $q = (2n)m + k = (2 \times 15) \times 2 + 41 = 101$), which is the same size of Piczak-CNN, 83% for segments of size 61 frames and 82% for segments of size 86 frames. The segment size is controlled by the extra frames $k$ with the rest of the model parameters unchanged. Moreover, to avoid the influence of the intermediate data representation on the MCLNN reported accuracies, we adopted the same spectrogram representation (60 Mel + Delta) used by the Piczak-CNN.

**Fig. 6.** Confusion matrix for the ESC-10 dataset. Classes: Dog Bark(DB), Rain(Ra), Sea Waves(SW), Baby Cry(BC), Clock Tick(CT), Person Sneeze(PS), Helicopter(He), Chainsaw(Ch), Rooster(Ro) and Fire Cracking(FC)

In a different evaluation to the MCLNN against the CLNN, we used the same architecture of the MCLNN, but without the mask to benchmark the CLNN. The highest CLNN accuracy is 77.5% at $k$=25, and the lowest one at $k$=1 is 73.25%, both are higher than the work reported in [26]. The MCLNN demonstrate accuracies that surpass the ones of the CLNN with the same architectures and segment sizes, which emphasizes the important role of the mask due to the properties discussed earlier.

Fig. 6 shows the confusion across the ESC-10 classes. The Clock tick sound is confused with fire cracking and dog barks due to the nature of the short duration of this category. There is also high confusion among the Chainsaw sound and both the Rain and Sea Waves due to the common low tones across these classes. The Baby Cry sound was detected with an accuracy of 97.5% and the lowest accuracy of 65% was for the Clock Ticks due to its short event duration.

## 6    Conclusions and Future Work

In this work, we have explored different experimental settings for the ConditionaL Neural Network (CLNN) designed for multidimensional temporal signal recognition and the Masked Conditional Neural Network (MCLNN) that extends upon the CLNN by enforcing a systematic binary mask. The mask plays the role of a filterbank embedded within the network, and it automates the handcrafting of the features through considering a different combination of features concurrently. We have used the MCLNN for environmental sound recognition using ESC-10 dataset, and it achieved accuracies that surpassed state-of-the-art convolutional neural networks using fewer parameters. Future work will consider different order $n$ across the layers and deeper MCLNN architectures. We will also consider applying the MCLNN to temporal signals other than sound.